\documentclass[sigconf]{acmart}
\usepackage{times}
\usepackage{latexsym}

\usepackage{subfigure}
\usepackage{graphicx}
\usepackage{amsmath}
\usepackage{enumerate}
\usepackage{enumitem}
\usepackage{multirow}
\usepackage{multicol}
\usepackage{balance}
\usepackage{color, colortbl}
\AtBeginDocument{%
  \providecommand\BibTeX{{%
    \normalfont B\kern-0.5em{\scshape i\kern-0.25em b}\kern-0.8em\TeX}}}

\setcopyright{acmcopyright}
\copyrightyear{2020}
\acmYear{2020}
\newcommand{\bert}{{\texttt{BERT}}~}
\newcommand{\distilbert}{{\texttt{DistilBERT}}~}
\newcommand{\bertlarge}{{\texttt{BERT-Large}}~}

\newcommand{\pool}{\texttt{Pool}}

\newcommand{\softmax}{\texttt{softmax}}
\newcommand{\cse}{\texttt{cross\_entropy}}

\newcommand{\hide}[1]{}
\newcommand{\zc}[1]{{\color{blue}[#1]}}

\newcommand{\cls}{[\mathrm{CLS}]}
\newcommand{\sep}{[\mathrm{SEP}]}

\newcommand{\vx}{\vec{x}}

\newcommand{\bs}{\mathbf{s}}
\newcommand{\bW}{\mathbf{W}}
\newcommand{\reals}{\mathbb{R}}

\newcommand{\tpm}{\(\pm\)}
\newcommand{\std}[1]{\((\pm#1)\)}

\newcommand{\upward}{\(\uparrow\)}
\newcommand{\downward}{\(\downarrow\)}

\newcommand\vpara[1]{\par\noindent\textbf{#1}\ }

\definecolor{LightCyan}{rgb}{0.88,1,1}

\begin{document}

%%
%% The "title" command has an optional parameter,
%% allowing the author to define a "short title" to be used in page headers.
\title{Probing and Fine-tuning Reading Comprehension Models for Few-shot Event Extraction}

%%
%% The "author" command and its associated commands are used to define
%% the authors and their affiliations.
%% Of note is the shared affiliation of the first two authors, and the
%% "authornote" and "authornotemark" commands
%% used to denote shared contribution to the research.
\author{Rui Feng}
\authornote{Equal contribution.}
\email{rfeng@gatech.edu}
\affiliation{%
  \institution{Georgia Institute of Technology}
}

\author{Jie Yuan}
\authornotemark[1]
\email{yuanj17@mails.tsinghua.edu.cn}
\affiliation{%
  \institution{Tsinghua University}
}

\author{Chao Zhang}
\email{chaozhang@gatech.edu}
\affiliation{%
  \institution{Georgia Institute of Technology}
}

%%
%% By default, the full list of authors will be used in the page
%% headers. Often, this list is too long, and will overlap
%% other information printed in the page headers. This command allows
%% the author to define a more concise list
%% of authors' names for this purpose.

%%
%% The abstract is a short summary of the work to be presented in the
%% article.
\begin{abstract}

    We study the problem of event extraction from text data, which requires both detecting target event types and their arguments. Typically, both the event
    detection and argument detection subtasks are formulated as supervised sequence labeling problems.
    We argue that the event extraction models so trained
    are inherently label-hungry, and can generalize poorly across domains and text genres.
    We propose a reading comprehension framework for event extraction. Specifically, we formulate event detection as a textual entailment prediction
    problem, and argument detection as a  question answering problem.
    By constructing proper query templates,
    our approach can effectively distill rich knowledge
    about tasks and label semantics
    from pretrained reading comprehension models. Moreover, our model can be fine-tuned with a small amount of data to boost its performance.
    Our experiment results show that
    our method performs strongly for zero-shot and few-shot event extraction, and it achieves state-of-the-art performance on the ACE
    2005 benchmark when trained with full supervision.

\end{abstract}

%%
%% The code below is generated by the tool at http://dl.acm.org/ccs.cfm.
%% Please copy and paste the code instead of the example below.
%%

%%
%% Keywords. The author(s) should pick words that accurately describe
%% the work being presented. Separate the keywords with commas.
%\keywords{datasets, neural networks, gaze detection, text tagging \zc{during submission, you do not need keywords, right?}}

%% A "teaser" image appears between the author and affiliation
%% information and the body of the document, and typically spans the
%% page.

%%
%% This command processes the author and affiliation and title
%% information and builds the first part of the formatted document.
\maketitle
\section{Introduction}

Event extraction is one of the most important and challenging tasks in information extraction.
Event extraction consists of two subtasks: 1) The first task, event detection, is
to detect if natural language text describes the
occurrence of certain events. 2)
The second task, argument detection, aims to find the attributes and participants, such as ``when'', and ``where'', and  ``who'', to the events.
Typically, both tasks are formulated as \textit{supervised sequence labeling} problems:
event detection is usually formulated as detecting the trigger words or phrases that best ``indicate'' an event;
and  argument detection is formulated as identifying entities that serve as arguments, such as ``who'', ``when'', and ``where'' for that event type.
Various sequence labeling models~\cite{yang2019exploring,orr2018event,hong2011using,liao2010using,chen2015event,chen2017automatically,nguyen2015event,nguyen2016joint,wang2019adversarial,sha2019jointly,yang2016hierarchical,mehta2019event,nguyen2018graph,lu2019distilling,du2020event}
have been proposed, which can be trained
on an annotated corpus and then used for recognizing event triggers and arguments at test time.

For example, the following sentence is extracted from ACE05 event extraction benchmark~\cite{doddington2004automatic}:
\begin{quotation}
   ``Orders went out today to \underline{deploy} 17,000 US soldiers in the \textit{Persian Gulf region.} ''
\end{quotation}
In the above example, an event ``Transport'' happened, \underline{deploy} is annotated as its trigger word, and \textit{Persian Gulf region} is the
\textit{Destination} argument to the event, and \textit{17,000 US soliders} would be the \textit{Artifact} (interpreted as passengers). 

However, formulating event extraction as supervised sequence labeling tasks have several drawbacks.
For event detection,
the annotation of event trigger words is of high variance. As the definition of ``trigger words'' is the word that most ``clearly'' expresses the occurrence of the event~\cite{ace05guide,liu2019event}, it is inherently noisy and time-consuming to label, especially in complex documents~\cite{liu2019event}.
This requires developing a specific set of rules governing ambiguity during annotation. Even so, the model is not necessarily able to recognize, or benefited from, such knowledge.
The annotation of arguments suffers from the same problem, as the span can be arbitrary (is it ``U.S. soldiers'' or ``17,000 U.S. soldiers'', or just simply ``soldiers''? All of them are
valid answers.)
%For example, in the following example: ``\underline{Former} senior banker Callum McCarthy begins what is one of the most important jobs...'', ``\underline{Former}'' is considered the trigger of the event that McCarthy quit his job as a banker.
%However, ``former''
% \zc{better to give an example here}
%First, the trained model suffer from high variance in data annotation, especially in event trigger words. The definition of trigger words can be ambiguous and the words on its own is not necessarily correlated to the semantics of events. Hence, many previous works need to manually introduce structural features, such as dependency parse, to improve span detection~\reminder{ref}.
Some existing approaches~\cite{liao2010using,sha2019jointly,duan2017exploiting,yan2019event,liu2018jointly}
attempt to resolve these issues by introducing more complex structural features (such as dependency parses and document-level information) or use more complex neural sequential
models.
But as a result, learning such complex models becomes highly label hungry, even when powerful pretrained language models are used~\cite{wang2019adversarial,yang2019exploring,du2020event}.
Furthermore, the learned model can easily overfit and be vulnerable to domain shift.
%As observed in ~\cite{orr2018event}, almost all traditional works
%suffer from severe  performance drop, compared with reported in paper, by a large margin when rigorously tested. \zc{this sentence is hard to
%  understand, need to rewrite and explain harder.}
As reported in \cite{orr2018event}, when rigorously tested with multiple random initializations, many works suffer from a severe performance drop compared with the best performance reported in their papers.

We propose to formulate event extraction as a machine reading comprehension task.
The general format of machine reading comprehension is that given a \emph{query} and a context sentence, the algorithm finds the answer to the query conditioned on the context sentence.
Specifically, we formulate event detection as a textual entailment prediction task. \hide{, where given each sentence as the context,
the model predicts if an event occurs by predicting if a description of the event is entailed by the sentence, without detecting or using the event's trigger word.}
The underlying intuition is that, suppose a sentence describes an event, then a statement that this event has happened would be a natural entailment to the first sentence.
Compared with formulating event extraction as a sequence labeling problem,
the benefit of such an entailment formulation is twofold.
First,
the model can  distill knowledge from pretrained language models and be appealing for few-shot or even zero-shot settings. Second, it can avoid
having to pinpoint event triggers and be able to flexibly exploit complex event semantics in the sentence.

% \zc{do not just say how you do it, explain the key ideas/insights behind your approach. what's the benefits of formulating event detection as TE? And
%   how does it address the drawbacks you mentioned earlier? Some important points should also be mentioned here, for example, distilling knowledge from
% pre-trained reading comprehension model (in zero shot settings), and labeled data can be transformed to train the reading comprehension model.}

\begin{figure}[t]
  \includegraphics[width=0.9\linewidth]{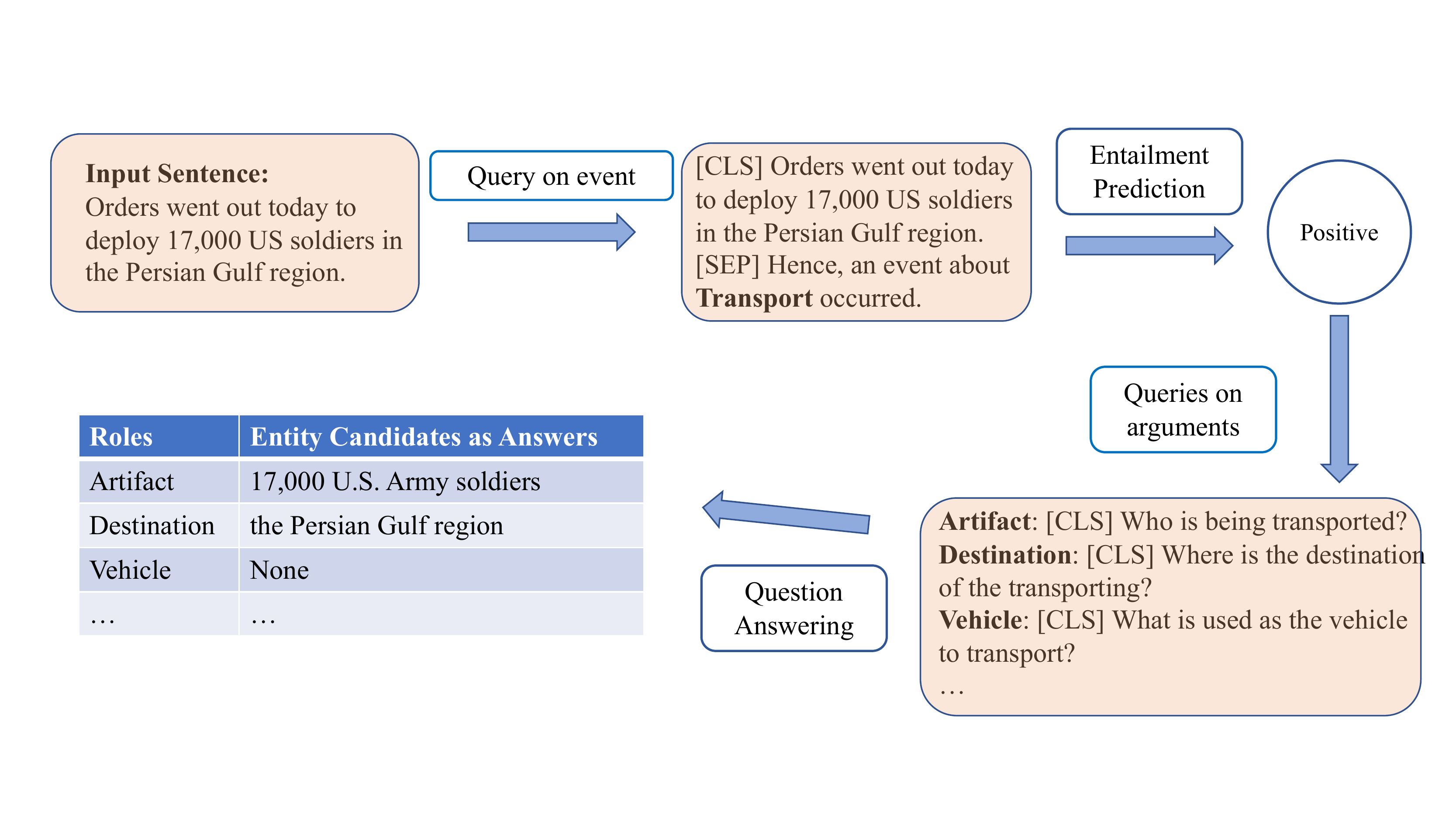}
  \caption{Illustration of our framework. 
  Given an input sentence, our framework predicts an event by predicting if a statement about the event is a logical entailment to the sentence with a textual entailment prediction module. 
  To extract arguments, we construct natural language questions about the argument and extract the answer with a question answering module from the sentence. 
  }
  \label{fig:illustration}
\end{figure}

We formulate argument extraction as a question answering problem. Finding an argument to an event can be naturally posed as finding an answer to a
question like ``Who was the attacker?'' or ``Where is the destination of the transport?''. 
In principle, argument extraction can be formulated as a textual entailment problem as well, but this requires generating a statement for each event-argument-entity pair. This would require pre-identify entities and the training could be inefficient. 
%\zc{readers may have a question here: why not formulate it as entailment again?}
By reframing argument extraction as QA, we can again transfer knowledge from pretrained models on how to find answers from text data.
In Section~\ref{sec:method}, we give detailed descriptions on how to pose event extraction tasks as reading comprehension tasks, and how to solve them.
The illustration of our model can be found in Figure~\ref{fig:illustration}.
%We construct question given events and argument types from either predefined templates or the descriptions in the annotation guide. \zc{similar
%  problem here, the description of our method is too weak. Need to justify and sell.}

We conducted experiments and analyzed model performances in zero-shot, few-shot, and fully-supervised settings in Section~\ref{sec:exp}. Our major
findings from the experiments are: (1) Our proposed approach of formulating event extraction as reading comprehension can
effectively distill knowledge by probing pre-trained models, and
achieve strong performance for zero-shot event detection without training data. (2) The  performance of our model increases largely when the model is
fed a small amount of labeled data. In fully-supervised settings, it achieve state-of-the-art performance. (3) We found that trigger words do not
improve performance for either event detection or argument detection, which justifies our design of discarding trigger words for event detection.

The major contributions of this paper are as follows:
\begin{enumerate}
    \item We propose a reading comprehension framework for event and argument detection. Comparing with existing works, our framework is more capable in
      exploiting pre-trained reading comprehension models on tasks and event semantics.
    \item We are the first to propose a reading comprehension framework of event detection and argument detection without trigger words.
    \item
By probing and fine-tuning
      pretrained reading comprehension models,
      our approach achieves much stronger performance for
 low-shot event extraction compared with the baselines; and it
 achieves
 state-of-the-art
performance for fully-supervised event detection on the ACE 2005 benchmark.
\end{enumerate}

\section{Preliminary}

% This is outdated.
% % structure of this section:
% %     1. tasks
% %         - event extraction tasks: input and output, objectives.
% %         - MRC tasks: inputs, outputs, and conversion from ee to mrc.
% %         Immediately following this should be query generation as the reader would be curious in this crucial piece of information on how queries should be generated.
% %     2. query generation. templates, descriptions, and guidebook.
% %     3. model details.
% %         - intro to bert.
% %         - and how queries described in previous parts are fed into model.
% %         - how to predict, and loss function.

% Current structure:
% 1. event detection tasks.
% 2. reading comprehension tasks.
% 3. \bert model.

\subsection{Tasks Definition}
\label{sec:tasks}
\vpara{Event extraction.}
The event extraction task consists of two subtasks: event detection and argument detection. We describe these tasks as follows.
\begin{itemize}
    \item \textit{Event Detection.}
        Following ~\cite{liu2019event}, we formulate event detection as a multi-class sentence classification problem.
        Each input sentence \(X=[x_1,\ldots,x_n]\) is associated with a binary label vector \(y\in\reals^m\) where \(m\) is the number of event types. Each component corresponds to an event type and is \(1\) if and only if the input sentence describes that event.
        Our goal is to predict for each event whether it is described in the given sentence.
    \item \textit{Argument detection.}
        Following most related works, e.g. ~\cite{chen2015event}, each sentence is annotated \emph{entities} as candidate arguments. Argument
        detection aims to determine, given the sentence, a described event and an entity, whether the entity serves as one of the predefined argument
        roles to that  event.
\end{itemize}

\vpara{Machine Reading Comprehension.}
%  intro paragraph
\hide{
In this paper, we provide a working definition of machine reading comprehension tasks (MRC) as, given a context sentence \(X\), we feed the machine, or
model, a query \(Q\), and it is supposed to return an answer to \(Q\). \zc{this sentence is too fragmented.}
% intro to two mrc tasks, and how/why event detection and argument detection are transformed to them.
}
%In this paper, we provide a working definition of machine reading comprehension (MRC) tasks.
We here give a working definition of machine reading comprehension (MRC) tasks.
Given a context sentence \(X\), we feed the machine (model) a query \(Q\) in form of natual language sentences. The model is expected to return the answer to \(Q\) conditioned on the context \(X\).

Specifically, we consider two MRC tasks: 1) textual entailment prediction and 2) question answering.

\begin{enumerate}
    \item \textit{Textual Entailment Prediction (TE).}
    Textual entailment prediction, as the name suggests, aims at classifying if a sentence is the entailment of another.
    Specifically, given sentence \(X\) and \(Q\), we query the model if \(Q\) is contradictory to, entailed by, or neutral to \(X\).
    Recognizing textual entailment can be formulated as a three-way classification task over sentence pair \((X, Q)\).
    \item \textit{Question Answering (QA).} We consider span-selection based question answering. In other words, we assume that the answer to the question can be found as a span of words in a given piece of text.
    Specifically, given sentence \(X\) and question \(Q\), the model finds a span in \(X\) by predicting the starting and ending token of the span and returns it as the answer.
\end{enumerate}
\section{Our Approach}
\label{sec:method}
In this section, we introduce our framework to solve event extraction tasks by formulating them into reading comprehension problems. We cast
event detection as a textual entailment prediction problem based on the intuition that a sentence should entail an event if the latter is described in
that sentence. We cast argument detection as a question answering problem since questions can be naturally asked about the specific arguments of
an event. In the following, we first provide high-level descriptions of such a framework in Section \ref{sec:method:event} and Section \ref{sec:method:arg}. We
describe how we generate queries for these two subtasks in Section \ref{sec:query_generation}, and finally detail our model along with its probing and
training procedures in Section \ref{sec:method:bert_backbone}.

\subsection{Event Detection as Textual Entailment}
\label{sec:method:event}
To pose event detection as textual entailment prediction, given sentence \(X\) and an event type,
we first construct a statement \(Q\) that claims the event type has happened. The task of determining whether an event has
happened in \(X\) is then translated to judging if \(Q\) is a natural entailment to \(X\). For example:
\begin{quotation}
  (\(X\)) David is leaving to become chairman of London School of Economics. (\(Q\)) \textit{Hence, an event about \texttt{Start of Position} occurred.}
\end{quotation}
In the above, the first sentence is the original input sentence to event detection, and the second italic sentence is a statement constructed for the queried event type, \texttt{Start-Position}.
If the statement is entailed by the first sentence, we predict that the event has happened.
We will describe in detail how to construct such \(Q\) statements in Section~\ref{sec:query_generation}.

\subsection{Argument Detection as Question Answering}
\label{sec:method:arg}
To convert argument detection to question answering, we construct a ``Wh''-question \(Q\) (a question that starts with an interrogative word, such as ``What'', ``Who'', and 'Where', etc.) about the concerned event and argument type, like the following:
\begin{quotation}
    \textit{
      (\(Q\))   Where did the \texttt{Meeting} take place?
    }
    (\(X\))    But the Saint Petersburg summit ended without any formal declaration on Iraq.
\end{quotation}
Where the first italic sentence is a question about event \texttt{Meet} and the queried argument type is ``Place''.
We construct questions from fixed templates as well as manually written question forms for each possible combination of events and argument types. See Section~\ref{sec:query_generation} for details.

\subsection{Query Generation}
\label{sec:query_generation}
% connecting to the previous section

We introduced event extraction and machine reading comprehension tasks
in Section~\ref{sec:tasks},
and how can the former be transformed as the latter in Section~\ref{sec:method:event},~\ref{sec:method:arg}.  
In this section, we describe how to generate queries for the two event extraction tasks, and how are combined with input sentences. 
After this, all preparations would be made for the model computation flow. 

It is of crucial importance 
%\zc{why crucial? Need to explain. for example, you want to better probe/distill knowledge from
%  pre-trained reading comprehension models. Then you have not mentioned you will use reading comprehension models in this section, you need to
%  say that.}
to generate high quality queries, i.e. statements about events for entailment-prediction-based event detection, and questions about events and argument types for question-answering-based argument detection, because the quality of queries determines 1) how the model distills information about tasks and labels from pretrained weights, and 2) how the model connect the semantics of events, arguments, and the sentence context. 

\vpara{Statements for events.}
%For each event type, we first process the event label name, resulting in a noun or noun phrase.
We assume that each event type has a label name. To construct a statement about an event, we simply fill in the label name in the following template:
\begin{quotation}
    Hence, an event about \texttt{[EVENT]} happened.
\end{quotation}
where \texttt{[EVENT]} is the placeholder for event names.

This statement is supposed to serve as a guide to distill knowledge from the language model about both the \emph{task} and the \emph{label}.
For the  \emph{task}, i.e. event detection,  an ideal model should be informed that it is supposed to recognize ``what has happened''.
For the  \emph{label}, the model should recognize clues on the semantics of the event by its label name.
In addition to label name, we expect that an natural langauge description of the event would further help us distill knowledge from the model.
So, we append an optional piece of description on the event, acquired from the data's annotation guide~\cite{ace05guide}, to the above statement. Some examples of event descriptions are given in Table~\ref{tab:event_desc}.

\begin{table}[h]
    \centering
    \begin{tabular}{c|p{60mm}}
    Event & Description \\ \hline
       Be-Born  & \textit{A \textbf{Be-Born} Event occurs whenever a PERSON Entity is given birth to.} Please note that we do not include the birth of other things or ideas. \\ \hline
       Marry & \textit{\textbf{Marry} events are official Events, where two people are married under the legal definition.} \\ \hline
       Divorce & \textit{\textbf{A Divorce} event occurs whenever two people are officially divorced under the legal definition of divorce. We do not include separations or church annulments. }\\ \hline
      Transfer-Money & \textit{\textbf{Transfer-Money} events refer to the giving, receiving, borrowing, or lending money when it is not in the context of purchasing something.} The canonical examples are: (1) people giving money to organizations (and getting nothing tangible in return); and (2) organizations lending money to people or other orgs.  \\
    \end{tabular}
    \caption{Example event descriptions.}
    \label{tab:event_desc}
\end{table}

\vpara{Questions for events and argument types.}
Similar to events, each argument type has a label name as well.
We have two options to construct a question for a pair of event and argument.
First and most straightforwardly,
we could use a fixed template similar to event statement:
\begin{quotation}
    Who or what participated as role
    \texttt{[ARGUMENT]} in the event \texttt{[EVENT]}?
\end{quotation}
where argument names and event names are filled in respective slots.

This rather inflexible approach does not provide much information on the relation between events and arguments,
since we assume the same query structure between all event-argument paris. It is more natural to ask questions differently, specific to the concerned event and argument. For example, ``What is the Person in event Start-Position?'' is a lot less natural than ``Who started a new position?''.
So, we manually composed a question for every pair of events and arguments based on descriptions in the annotation guide~\cite{ace05guide}. We do so by converting the description text to a question with minimal edit while ensuring that the concerned argument type appears in the question. For example, from ``The people who are married.'' we construct ``Who are the married person?''. ``People'' are changed to ``Person'' to match the concerned argument type name. 
Examples
are given in Table~\ref{tab:arg_questions}. 
\begin{table}[h]
    \centering
    \begin{tabular}{c|c|p{50mm}}
    Event & Arg. Type & Question \\ \hline
       \multirow{2}{*}{Marry} & Person & Who are the married person? \\
       & Where & Where does the marriage take place?\\ \hline
       \multirow{2}{*}{Attack} & Attacker & Who is the attacker? \\
       & Target & Who is attacked?
    \end{tabular}
    \caption{Example questions for event-argument pairs. }
    \label{tab:arg_questions}
\end{table}

\subsection{Model Details}
\label{sec:method:bert_backbone}
In the previous section we described how to transform event extraction tasks to reading comprehension tasks.
In this section, we detail our model for solving these tasks, as well as how the model works in zero-shot and supervised settings.

% Structure of this subsection:
% 1. BERT base model.
% 2. BERT for two reading comprehension tasks.
\vpara{\bert masked language model. }
First, we give an introduction to the backbone \bert model that outputs hidden representations for tokens and sentences as the backbone for textual entailment prediction and question answering.

\bert~\cite{devlin2018bert_} model is a masked language model that consists of deep multihead attention layers.
Given a pair of input sentences \(X^{(1)}, X^{(2)}\), \bert first runs WordPiece~\cite{schuster2012wordpiece} tokenizer to tokenize the both sentences into a sequence of token ids: \( [ x_{1}^{(i)}, \ldots, x_{n_i}^{(i)} ], i=1,2 \). Then, two sentences are concatenated into one sequence:
\begin{equation}
    S = [\cls, x^{(1)}_1,\ldots, x^{(1)}_{n_1}, \sep, x_1^{(2)}, \ldots, x^{(2)}_{n_{2}}, \sep]
\end{equation}
where \(\cls\) is a special token at the beginning that is supposed to aggregate information in the two sentences, \(\sep\) is another special token used to inform separation between sentences. This combined sequence of tokens is the input to \bert model.

\vpara{Textual entailment for event detection.}
To perform textual entailment prediction,
we attach a linear classifier on top of the sentence embedding.
To obtain sentence embedding, we use \bert's default pooling method:
\begin{equation}
    \vx = \pool(\bert([X, Q]))
    \label{eqn:bert_sentence_embed}
\end{equation}
where \(\pool\) operator just extracts the output embedding of the \(\cls\) token.

A linear layer transforms the sentence embedding to logits:
\begin{equation}
    [l_0, l_1, l_2] = \vx^T \bW_E
\end{equation}
where \(\bW_E\in\reals^{d\times 3}\), \(d\) is the dimension of the hidden spaces, and \(l_0,l_1,l_2\) are logits for predicting contraditory, entailment, and neutral, respectively. We consider that the underlying event to happen if and only if it predicts entailment. Hence,
the final scores for predicting the described event in the statement is obtained by a softmax function:
\begin{equation}
    [p_0, p_1] = \softmax(l_0+l_2, l_1)
\end{equation}
The final loss function is the cross entropy function:
\begin{equation}
    \ell_E = \cse(p_0, p_1; y) = -((1-y)\log p_0 + y \log p_1)
    \label{eqn:entail_loss}
\end{equation}
where \(y\) is a binary label that indicates whether the described event happens.

\vpara{Question answering for argument detection.}
Given a question about an event and argument type, we need to predict the probability for each token as the starting or ending of the answer span.
First, we collect the output embeddings from \bert for each token in the original sentence \(X\):
\begin{equation}
    [\vx_1,\ldots,\vx_n] = \bert(S)
\end{equation}
To obtain possible spans, we use two linear classifiers on token embeddings:
\begin{equation}
    \begin{aligned}\
    [\bs^l_1,\ldots, \bs^l_n] & = \softmax([\vx_1,\ldots, \vx_n]^T \bW_l) \\
    [\bs^r_1,\ldots, \bs^r_n] & = \softmax([\vx_1,\ldots, \vx_n]^T \bW_r)
    \end{aligned}
\end{equation}
where \(\bW_l, \bW_r \in \reals^{d\times 1} \), \texttt{softmax} normalizes logits across tokens, and \(s_i^l, s_i^r\) are scores of the \(i\)th token being the start or end of the answer span.
Given ground-truth labels of span starts \(y_i^l\) and ends \(y_i^r\),
the model optimizes the cross entropy loss:
\begin{equation}
    \begin{aligned}
        \ell_Q & = \ell_Q^l + \ell_Q^r \\
        \ell_Q^l & = - \sum_{i=1}^n y_i^l \log s_i^l \\
        \ell_Q^r & = - \sum_{i=1}^n y_i^r \log s_i^r
    \end{aligned}
    \label{eqn:qa_loss}
\end{equation}
to make a prediction,
we follow ~\cite{devlin2018bert_} and
predicts the span as \(i, j\) with the highest \(s_i^l + s_j^r\) satisfying \(i<j\).
Additionally, unlike in the QA setting in SQuAD~\cite{rajpurkar2016squad} and adopted by \bert~\cite{devlin2018bert_},
not every queried event and argument type pair has an answer.
Hence, we only predict the answer when both \(s_i^l\) and \(s_j^r\) are greater than a certain threshold, typically \(0.5\).
\section{Experiments}
\label{sec:exp}
We report in this section the experiemnt results on event detection.
In Section~\ref{sec:data} we briefly introduce the dataset, ACE 2005, that we experiment on. 
For both of the two tasks, 
first, we probe the pretrained language model's ability to infer events without specific training in Section~\ref{sec:event:zero_shot},~\ref{sec:arg:zero_shot}.
Then, in Section~\ref{sec:event:sup},\ref{sec:arg:sup}, we report results when the model is finetuned on event detection and argument detection.
Ablation studies were reported separately in Section~\ref{sec:event:ablation},~\ref{sec:arg:ablation}.

\subsection{Data}
\label{sec:data}
We use ACE05 Multilingual Training Corpus~\cite{doddington2004automatic} for event detection and argument detection. 
It contains 33 event types, 28 argument types, and sentences that come with various documents. 

ACE2005 is observed to have domian shifts between its official training, development, and test set split: the training and dev. set contain informal documents such as web logs, while the test set is a collection of newswire articles~\cite{doddington2004automatic,orr2018event}. 

We follow the data split used by ~\cite{chen2015event}. Table~\ref{tab:data_stat} gives data statistics. 

\begin{table}
    \begin{tabular}{l|ccc}
       Data split & \# sentences & \# events & \# arguments \\ \hline
       Train &  14626 & 4309 & 7702 \\ 
       Dev & 870 & 492 & 923 \\ 
       Test & 708 & 422 & 887 
    \end{tabular}
    \caption{Data statistics for training, development, and test set. We list here number of sentences, events, and arguments. }
    \label{tab:data_stat}
\end{table}

\subsection{Event Detection Experiments}
\label{sec:event_detection}
\hide{For event detection, we use \bert and \distilbert models finetuned on MNLI. Since we could not find a pretrained \bertlarge on MNLI and finetuning it exceeds overburden our computation resources, we do not experiemnt with \bertlarge. }

\begin{table}[tbp]
    \small
    \centering
    \begin{tabular}{c|ccccc}
        & Precision~\upward & Recall~\upward & F1~\upward & Threshold & p-value~\downward \\ \hline
        Random & 5.75 & 86.28 & 10.79 & 0.157 & 0.63 \\
        MTP + TE & 12.50 & 29.09 & 17.40 & 0.825 & 0.00 \\ 
        MTP + QA & 15.13 & 17.05 & 16.03 & 0.849 & 0.00 \\
        QA & 1.03 & 100 & 2.03 & 0.0 & 0.00 \\
        \rowcolor{LightCyan}
        TE & 16.84 & 36.89 & \textbf{23.12} & 0.385 & 0.00 \\
    \end{tabular}
    \caption{Zero-shot learning for event detection. Precision, Recall, F1-scores are evaluated on test set, with Threshold chosen to maximize F1-score on the development set. Any instance with predicted probability greater than the Threshold is classified as positive. \(p\)-value indicates the the level of statistical significance that the model does separate ground-truth events from false events. \textit{Random} is the F1-score obtained by randomly generating scores from a uniform distribution on \([0, 1]\). }
    \label{tab:zero_shot_results}
\end{table}

\begin{table*}[tbp]
\begin{tabular}{p{0.25\textwidth}p{0.10\textwidth}p{0.25\textwidth}p{0.25\textwidth}}
 Sentence & Ground-truth & Top candidates from vocab & Top candidates from event names \\  \hline
    Jay Garner the retired general will go into Iraq soon with his troops soon. & \multirow{3}{0.10\textwidth}{Movement: Transport} & 
    just time to never what war has this today now & Attack (0.98) Injure (0.96) Transport (0.92) \\ \cline{1-4} 
     It would not have been necessary to fire those 17 people right away. & Personnel: End-Position & never just to time had christmas not certainly have now & Die (0.97) Injure (0.92) Trial-Hearing (0.91) \\ \cline{1-4}
     \hide{If he is willing to file a lawsuit on the basis of New York non-profit organization law, which might or might not be applicable to the USCF,}
     Why wouldn't he file a lawsuit on the basis of the USCF's violation of its own bylaws, which unquestionably ARE applicable to the USCF? & \multirow{4}*{Justice: Sue} & just never to has what not have already having had & Trial-Hearing (0.98) Sue (0.98) Injure (0.96) Arrest-Jail (0.90) Transfer-Money (0.85) \\ \hline
\end{tabular}
\caption{Examples of masked token predictions. We show the original sentence, ground-truth labels, top candidate for the [MASK] placeholder from both the entire vocabulary and only from the set of event names. For each top candidate event name, we also show their predicted probabilities. }
\label{table:mtp_examples}
\end{table*}

\begin{table*}[tbp]
    \centering
    \begin{tabular}{l|ccccccccc}
    \(K\)-shot & \(K=1\) & \(K=3\) & \(K=5\) & \(K=7\) & \(K=9\) \\
    \distilbert & 0.42~\std{0.29} & 0.42~\std{0.29} & 0.42~\std{0.29} & 0.42~\std{0.29} & 0.42~\std{0.29} s\\
        TED & \textbf{43.02} \std{2.36} & \textbf{51.60} \std{2.15} & 51.51 \std{2.59} & 56.20 (\tpm 3.11)  & 56.83 (\tpm 4.36) \\
        TED+D (1) & 23.10 (\tpm 5.20) & 45.94 (\tpm 4.80)  & 51.92 (\tpm 4.07) & 54.88 (\tpm 2.82) & 57.11 (\tpm 2.90) \\
        TED+D (5) & 17.80 (\tpm 4.29) & 45.76 (\tpm 5.17) & 52.44 (\tpm 2.47) & 54.68 (\tpm 3.55) & 59.62 (\tpm 2.63) \\
        %PQ-QA & 40.23 \std{1.30} & 48.88 \std{3.27} & 50.05 \std{1.40} & 54.59 \std{1.88} & 55.77 \std{3.83} \\
        %PQ-QA+D (1) & 11.81 \std{13.45} & 47.86 \std{2.27} & 49.72 \std{4.57} & 52.89 \std{3.36} & 54.40 \std{4.41} \\
        %PQ-QA+D (5) & 6.34 \std{13.09} & 42.60 \std{3.01} & 49.34 \std{2.86} & 52.48 \std{3.79} & 54.97 \std{2.38} 
        \hline
        \bert & 0.70~\std{0.40} & 0.70~\std{0.40} & 0.42~\std{0.29} & 0.42~\std{0.29} & 0.42~\std{0.29} \\ 
        TE & 36.00 \std{3.56} & 50.68 \std{1.51} & 53.31 \std{2.40} & 56.01 (\tpm 2.52)  & 57.46 (\tpm 3.24) \\
        TE+D (1) & 25.19 (\tpm 3.28) & 46.94 (\tpm 5.99)  & 53.96 (\tpm 2.96) & 55.33 (\tpm 2.99) & 59.56 (\tpm 1.91) \\
        TE+D (5) & 26.66 (\tpm 1.74) & 49.48 (\tpm 1.95) & \textbf{54.47} (\tpm 3.25) & \textbf{57.00} (\tpm 2.03) & \textbf{61.75}f (\tpm 2.12) \\
    \end{tabular}
    \caption{
        Results on few-shot event detection. We show mean scores and standard variances on micro F1-scores. 
        We tried both \distilbert and \bert as the backbone model. ``TED'' means pretrained \distilbert on textual entailment, and ``TE'' means pretrained \bert on the same task. 
        ``+D'' means trained with event description, and the number in the following parenthesis means the max number of used sentences in the description. 
    }
    \label{tab:few_shot_results}
\end{table*}

\begin{table}[tbp]
    \small
    \centering
    \begin{tabular}{l|ccc}
        Method & Precision & Recall & F1 \\  \hline
        DMCNN~\cite{chen2015event} & 75.6 & 63.6 & 69.1 \\ 
        Delta~\cite{lu2019distilling} & 67.30 & 69.62 & 68.44 \\ 
        VERB-QA~\cite{du2020event} & 71.12 & 73.70 & 72.39  \\
       % \bert~\cite{devlin2018bert_} & & & \\
        \hline
        \bert~\cite{devlin2018bert_} & 71.52 \std{0.19} &  70.48 \std{1.65} & 70.99 \std{0.82} \\ 
        Delta~\cite{lu2019distilling} & 70.97 & 70.78 & 70.88 \\ 
        DS-DMCNN~\cite{liu2019event} &  \textbf{75.7} & 66.0 & 70.5 \\ 
        \rowcolor{LightCyan}
        TE & 73.28 \std{2.13} & 76.29 \std{1.30} & \textbf{75.43} \std{1.48} \\
        TE-D (1) & 73.04 \std{3.21} & 75.82 \std{1.41} & 74.39 \std{2.29} \\ 
        TE-D (5) & {72.95} \std{3.03} & \textbf{78.20} \std{3.89} & {75.38} \std{0.53}  
    \end{tabular}
    \caption{Supervised results. The first group of methods are based on trigger detection, while the second group predicts events without triggers.}
    \label{table:supervised_eventsp}
\end{table}

\begin{figure*}[tbp]
    \centering
    \subfigure[Without description.]{
    \label{fig:distil_vs_bert:nodesc}
    \includegraphics[width=0.3\textwidth]{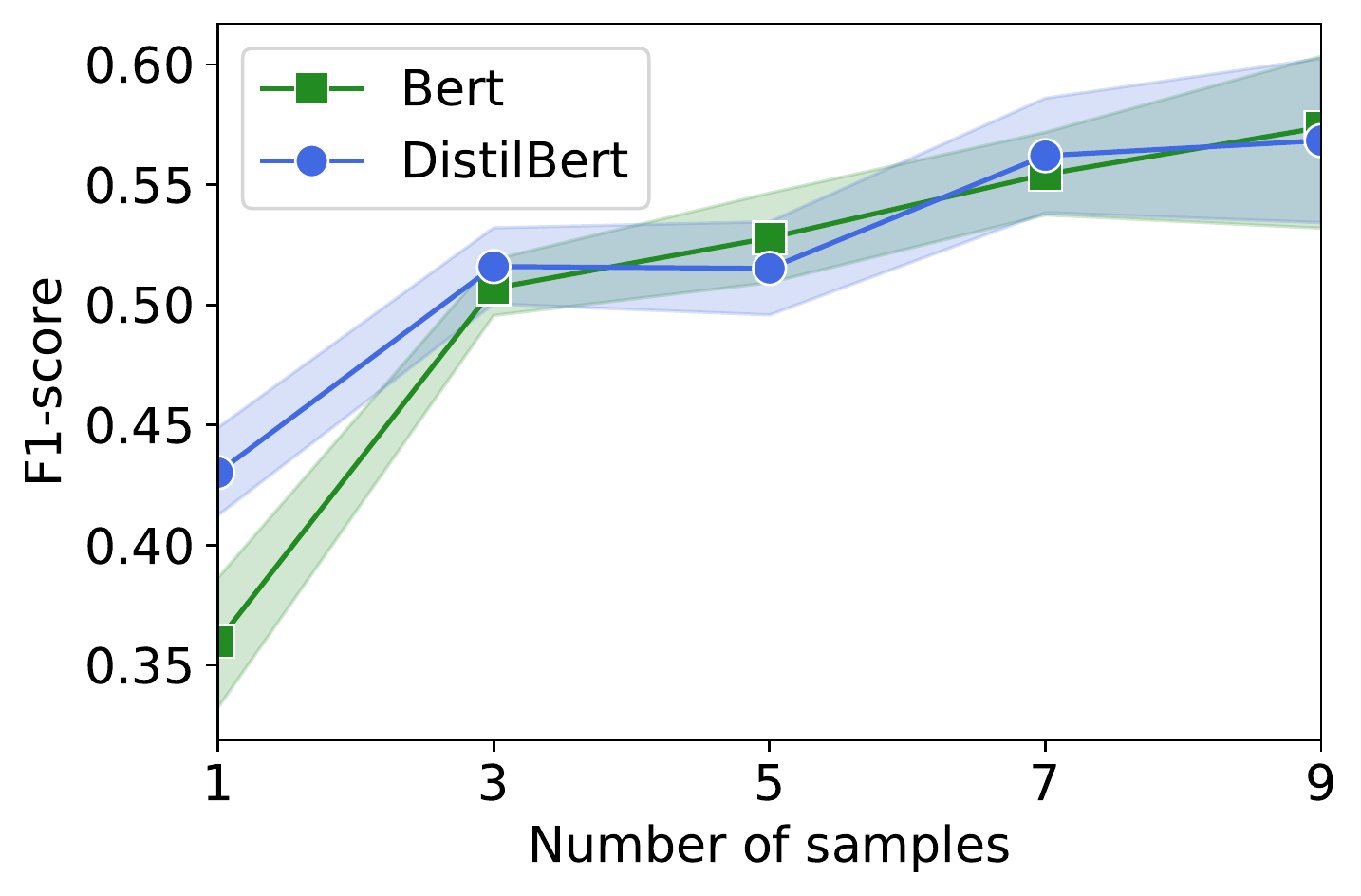}
    }
    \subfigure[With description, one sentence. ]{
    \label{fig:distil_vs_bert:desc1}
    \includegraphics[width=0.3\textwidth]{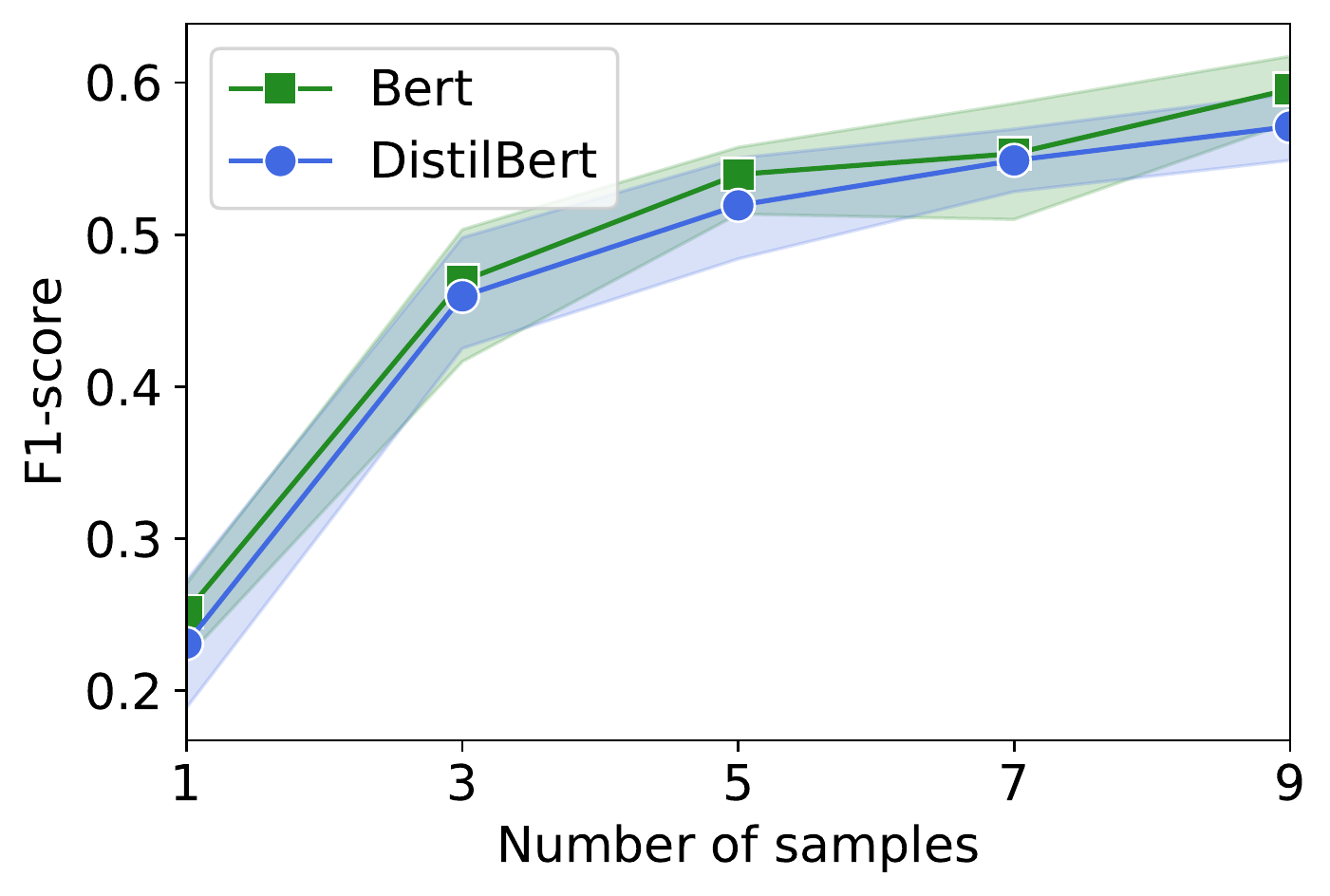}
    }
    \subfigure[With full description. ]{
    \label{fig:distil_vs_bert:desc5}
    \includegraphics[width=0.3\textwidth]{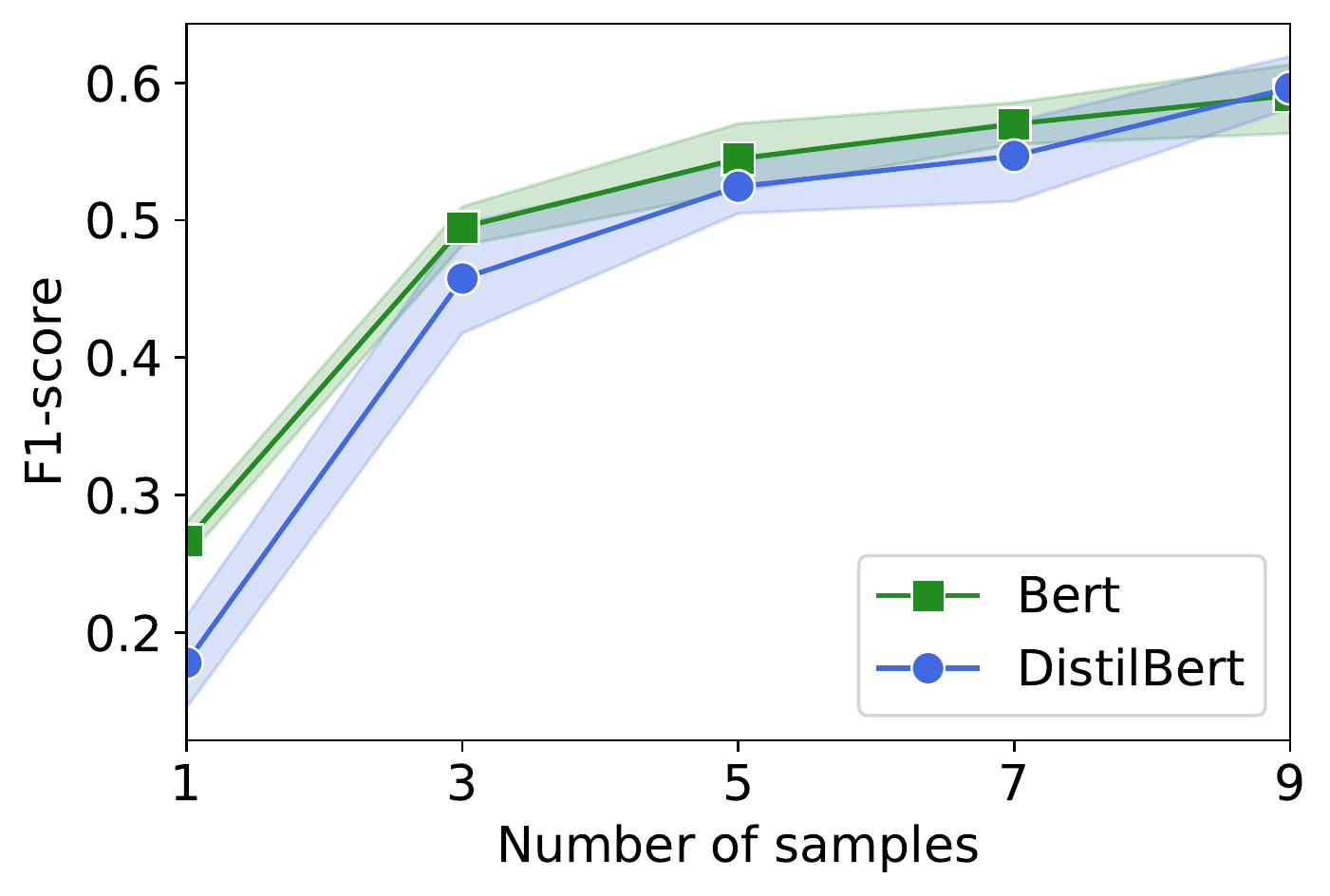}
    }
    \caption{\distilbert vs \bert on event detection in few-shot settings, with or without descriptions. Green lines with square marks are scores with \bert as backbone model, and blue lines with circle marks with are \distilbert. Strips represent confidence intervals. }
    \label{fig:distil_vs_bert}
\end{figure*}

\subsubsection{Probing without Supervision}
\label{sec:event:zero_shot}
We first experiment on predicing events without any training, but only exploiting potential knowledge in pretrained models alone. 

Concretely, we mostly experiment with \bert and the smaller \distilbert models pretrained on textual entailment prediction dataset, MNLI~\cite{mnli}\footnote{We used https://huggingface.co/textattack/bert-base-uncased-MNLI and https://huggingface.co/ishan/distilbert-base-uncased-mnli.}

We report performance on these following methods:
\begin{itemize}
    \item {Random}. This method simply generates random scores from uniform distribution on \([0, 1]\) as predictive probability as a refernce. 
    \item Textual entailment (TE). This is the our approach introduced in Section~\ref{sec:method:event} with \bert.
    \item Question answering (QA). We ask models question like ``What is the trigger for \texttt{[EVENT]}?'' and let the model do span detection of the trigger word, following the same method as in ~\ref{sec:method:arg}. We convert the prediction to sentence-level detection by predicing an event if any span with scores greater than threshold is predicted for that event. For this particularly model, we use \bert pretrained on SQuAD~\cite{rajpurkar2016squad}. 
    \item Masked token prediction (MTP). Instead of filling in event labels in query templates, we fill in \([MASK]\) special token as a placeholder and let \bert model predict what this token might be. 
    Events are classified based on the event label name's average score predicted by \bert to replace \([MASK]\). 
    For this model, the original pretrained \bert is used~\cite{devlin2018bert_}.
    The scores here are obtained by sigmoid function on individual logits, instead of softmax across entire vocabulary, to avoid arithmetic underflow. Specifically,
    \begin{itemize}
        \item \texttt{MTP+TE} predicts the \([MASK]\) token based on textual entailment formulation. 
        \item \texttt{MTP+QA} predicts the \([MASK]\) token based on a question answering formulation, where we attach a question like ``Did any event about \texttt{[MASK]} happen?'' before the input sentence. 
    \end{itemize}
\end{itemize}

In common binary classification settings, one predicts positive when the predicted probability is greater than \(0.5\). 
However, when the model is not trained the concerned data, although the model may contain certain bias towards the task, it is not necessarily \emph{calibrated} - i.e. it might predict favorably towards the right answer, but not necessarily to the extent that the predicted probability is always larger than \(0.5\). 
Hence, we manually choose a threshold \(p_0\), 
from 0 to 1 with step 0.01 that maximizes F1-score on the development set, and we use this threshold to separate predicted negative and positive class and report the performance on test set using this threshold.

Table~\ref{tab:zero_shot_results} shows the results for zero-shot event detection. We show the following metrics: Precision, Recall, F1, the optimal threhsold on the development set. 
In addition, for each of the methods, we performed a statistcal test on whether the distribution of scores on the ground-truth events is significantly different from the distribution of scores on false events. We use the standard Kolgomorov-Smirnov 2-sample test~\cite{massey1951kolmogorov,hodges1958significance} and report the \(p\)-value, the lower the better. 

From the results, we clearly see that entailment is a more natural formulation for \bert model as it contains more prior knowledge than question answering framework. If we use a question answering model to predict trigger span, the best performance is obtained when \(p_0=0.0\), effectively providing no useful information at all. 
We believe that it is because trigger words are not natural ``answers'' to questions, events are, but event names are not necessarily present in the text where events occur. Textual entailment does not rely on this, hence it is more able than question answering to solve event detection.

% case study.
In Table~\ref{table:mtp_examples} we showcase some examples of wrong predictions made by masked token prediction based on textual entailment. 
We see that when \(\bert\) doesn't predict the correct event as the one with the top probability, it still chooses the events that are semantically related to the sentence, although the corresponding event might not have actually happened. 

% construct a few better queries. See if the result gets better. 

\subsubsection{Few-shot Learning}
\label{sec:event:sup}
%Since that unlike in many other tasks, pretrained language models does not significantly outperform random guessing as explained in Section~\ref{sec:zero_shot}, 
Following zero-shot learning, we naturally want to test the model's ability to transfer knowledge to event detection given a few examples for each event type. 
We naturally want to test the model's ability to transfer to event detection task with a few learning examples and if the model can build upon a few samples the link between model's semantic knowledge and the reasoning of events. 

Table~\ref{tab:few_shot_results} shows experiemnt results on few-shot learning. 
We would like to highlight the following observations:
\begin{enumerate}
    \item \bert and \distilbert on their own is not capable of few-shot learning for our tasks, resulting in converging to trivial solutions. 
    \item \bert model does significantly outperform \distilbert. When \(K=1\), \distilbert works even better in textual entailment without description. 
    \item \bert's becames advantageous when the number of training data increases.
    \item Both \bert and \distilbert cannot efficiently utilize description information when \(K\) is small, resulting in a lower F1-score. When \(K\) increases, the performance gradually increases and matches the model without description. 
\end{enumerate}
Generally, in few-shot settings, using \(\distilbert\) could be a better choice, balancing performance and efficiency. 
As a comparison, \cite{peng2016event} achieved 67.6. They prefilled possible argument slots for candidate events using semantic role labeling tools, effectively summarizing event structures, and this piece of information is not readily available to \bert model. In the future, it would be an interesting direction to investigate if it is possible to use \bert model to more effectively extract event structures.

\hide{
In Table~\ref{tab:few_shot_results}, we show few-shot learning results of different models with or without event description. 
Event descriptions in the annotation guide~\cite{ace05guide} can contain as many as 5 sentences, exceeding the text length of the sentence greatly. 
This makes harnessing the information in description hard when the number of training data is small. 
Therefore, in addition to training with full event descriptions, we truncate event descriptions to contain their the first sentences in hope of making the model easier to learn how to use information from descriptions. 
We make the following observations. 
First, when the training samples are too small (from \(K=1\) to \(K=5\)), adding event descriptions undermines model performance, especially in by a large margin when \(K=1\). 
Furthermore, when training with only the first sentence in description, the performance drop is much lower. 
However, when we have more samples per class, the models with event descriptions gradually outperform those without, and models with full descriptions outperform with only one sentence. 
Furthermore, when trained without descriptions, one observes counter-intuitively that the standard variance across multiple experiments increase with the number of samples. 
}

\subsubsection{Fully Supervised.}
\paragraph{Baselines.}
Because the commonly used data for event detection, ACE05, is not freely available, there are only so few works that are open source whose code are complete and immediately runnable given data. This creates a problem for us to evaluate traditional methods in terms of sentence-level event detection performance.
Hence, for many of them, we could only report the results from the original paper. 

During our attempts to reproduce the experiments, event with official code, we still struggle to produce results consist with what was reported in papers. The same phenomenon was observed by ~\cite{orr2018event} where metrics can drop from 5 to 10 points in rigorously designed testing compared with reported results. 
Based on observations, we suspect such inconsistencies could be a result from lack of adherence to standard practices in terms of evaluation schemes, data processing, split of training/dev/test sets, and model selection, which all could influence the variance of validity of reported performance.

\hide{
It is worth noting that even when reproducing results for trigger-based detection, as is the setting of most previous works, we struggle to reproduce their reported results consistently using official code because of the inconsistencies and sometimes dubious practices we observed in terms of model selection, prediction and evaluation schemes, data processing, and the split of training/development/test sets, which all could influence the variance and validity of reported performance. 
Similar observation was made by ~\cite{orr2018event}, where they found that the performance of most previous works drop by a large margin, from 5 to even 10 points, in rigorous evaluation settings. 
We believe it is indeed necessary in this field to set standards on these matters
if any further progress in this field is to be made. }

Based on these reasons, we could only vouch for the validity of performances methods reported with standard variances, and the same observations and conclusion extends to argument detection as well. 
The following baselines are used as comparable sentence-level event detection methods without triggers: 
\begin{itemize}
    \item \bert~\cite{devlin2018bert_}. This based on the original pretrained \bert model, where event classification is done by learning a linear classifier on top of the pooled sentence embeddings. 
    \item Delta~\footnotemark~\cite{lu2019distilling}. This method is based on an adversarial framework where the model learns both discriminative and generalizable information. Although it is originally a trigger-based event detection method, we evaluate it here in terms of sentence-level scores. 
    \item DS-DMCNN~\cite{liu2019event}. This method performs event detection without trigger based on ~\cite{chen2015event}. 
\end{itemize}

The following baselines are trigger-based event detection methods, which we report here as a reference:
\begin{itemize}
    \item DMCNN~\cite{chen2015event}. This method proposed a dynamic pooling layer for CNN for event detection. 
    \item Delta\footnotemark[\value{footnote}]~\cite{lu2019distilling}. This is the same method as mentioned above. 
    \item VERB-QA~\cite{du2020event}. This is a QA based event detection framework where ``verb'' is used as a query to hint the model for trigger word detection and classification. 
\end{itemize}
\footnotetext{This method was originally proposed to detect and classify triggers. We report the performance of their method on both trigger-based event detection and sentence-level event detection, while the former is computed by ignoring if the span matches. The performance on this method is reported based on reproduction of their method of the authors' published code. The paper reported about 74.7 F1-score on trigger detection and classification.}

Our methods include TE, TE-D (1) and TE-D (5), which are the same as in the few-shot learning setting with \bert model. 
We observed significant performance drop with \distilbert, with only 70.04 averge F1 score. Therefore, we focus on performances on \bert model only, which can apparently utilize more data more effectively. 
From Table~\ref{table:supervised_eventsp}, our model achieves the best performance among baselines. 

\hide{
\begin{table}
    \centering
    \begin{tabular}{l|l|ccc}
        Setting & Method & Precision & Recall & F1 \\  \hline
        \multirow{4}{*}{w/ triggers} & DMCNN~\cite{chen2015event} & 75.6 & 63.6 & 69.1 \\ 
        & Delta~\cite{lu2019distilling} & 67.30 & 69.62 & 68.44 \\ 
        & VERB-QA~\cite{du2020event} & 71.12 & 73.70 & 72.39  \\
        & \bert~\cite{devlin2018bert_} & & & \\
        \hline
        \multirow{6}{*}{w/o triggers} & \bert~\cite{devlin2018bert_} & & & \\ 
        & DS-DMCNN~\cite{liu2019event} &  75.7 & 66.0 & 70.5 \\ 
        & Delta~\cite{lu2019distilling} & 70.97 & 70.78 & 70.88 \\ 
        & TE & & & \\
        & TE-D (1) & 73.04 \std{3.21} & 75.82 \std{1.41} & 74.39 \std{2.29} \\ 
        & TE-D (5) & 72.95 \std{3.03} & 78.20 \std{3.89} & 75.38 \std{0.53}  
    \end{tabular}
    \caption{Supervised results.}
    \label{tab:my_label}
\end{table}
}

\subsubsection{Discussions}
\label{sec:event:ablation}
\paragraph{Effect of query structure. }
In query templates we infused two kinds of knowledge: 1) information about the event detection task by providing a statement sentence, and 2) information about labels (event types) by filling in the statement event names. 
A natural question one would ask is if the statement structure is actually helpful or is it just the event names are enough.

\paragraph{Do event descriptions help?}
Based on experiments on few-shot (Table~\ref{tab:few_shot_results}) and fully supervised (Table~\ref{table:supervised_eventsp}) settings, 
we see that event descriptions can be a burden to model learning when the number of data is extremely small. While with increased data size and in the fully supervised setting, model with descriptions could have comparable perforance with the model without descriptions, it does not outperform the latter. 

However, event descriptions do provide valuable information on what are events and is sufficient on its own for a human annotator.  
This could suggest that \bert model cannot learn from descriptions more than it can learn from the data. This suggests that more work should be done on improving language models in understanding how to ``read'' a ``description'' or ``manual''. 

\paragraph{\bert vs. \distilbert}
In Figure~\ref{fig:distil_vs_bert} we compare the performance of \bert and \distilbert. 
It is clear from the figures that in few-shot settings, there is no statistically significant difference in performance, except for \(1\)-shot learning, where \distilbert is better without description and \bert, however, is better with descriptions. 
In fully supervised learning, the difference is significant where \(\distilbert\) achieves only 70 F1-score in average and \bert achieves 75. 
Therefore, when training labels are extremely scarce, it is sufficient to use \(\distilbert\) to effectively learn from these samples with less memory consumption and computational burden. 
In fully supervised learning, however, \bert's ability to utilize massive data is significantly better than \distilbert. 

\subsection{Argument Detection Experiments}
\label{sec:arg}
\subsubsection{Probing without Supervision}
\label{sec:arg:zero_shot}

Like in event detection experiments, we first probe pretrained \bert for QA tasks on SQuAD~\cite{rajpurkar2016squad} and see if they can predict arguments without training~\footnote{We used deepset/bert-large-uncased-whole-word-masking-squad2. }.
We experiment on the following query types:
\begin{enumerate}
    \item QA-Template. This constructs questions from templates in Section~\ref{sec:method:arg}
    \label{qa_exp_temp}
    \item QA-Guide. This constructs questions from descriptions in the annotation guide~\cite{ace05guide}.
    \label{qa_exp_guide}
    \item QA-Trigger. This constructs questions from a template similar to the one in Section~\ref{sec:method:arg}, but the question is asked about the trigger, like: ``What is the \texttt{[ARGUMENT]} in \texttt{[TRIGGER]}?''. %It is also essentially the same template used by ~\cite{du2020event}.
    \label{qa_exp_trigger}
    \item QA-Trigger-Plus. This is based on a similar template which includes both trigger word and event name. For example: 
    ``What is the \texttt{[ARGUMENT]} in event \texttt{[EVENT]} triggered by `\texttt{[TRIGGER]}'?''
    \label{qa_exp_trigger_plus}
\end{enumerate}
\ref{qa_exp_trigger} and ~\ref{qa_exp_trigger_plus} is the method used by ~\cite{du2020event}. Since we are doing event detection without triggers, it is necessary to include \ref{qa_exp_trigger} 
%and ~\ref{qa_exp_trigger_plus} 
as well to demonstrate the effect of removing trigger words on argument detection.

Table~\ref{tab:zero_shot_arg_results} shows results of models without training. Unlike in event detection, no threshold probability was manually chosen. We see that all four query templates perform similarly, and are all much better than random baseline. 
Also, we observe that questions with trigger words perform slightly better than those without. However, this is based on ground-truth annotation trigger words. 

In Table~\ref{tab:exp:arg_examples}, we show some examples of zero-shot predictions by generated queries.
Additionally, we manually wrote customized queries for each sentence based on the principle that the query should specify as much information as possible about the event, including other relevant arguments, like ``London's financial world'', and ''by the end of the year'', that are present in the input sentence. 

From Table~\ref{tab:exp:arg_examples}, we can observe that the custom query is much more semantically and contextually related to the input sentence. Compared with template-based queries which might ask ``Who started a new position'', or ``What is the artifact in Transfer-Ownership'', the written query is much more context specific. 
It provides stronger guidance for \bert model to extract the actual answer. 
Unfortunately, such queries are specific to sentence contexts and requires human writing, and it is hard to scale-up to the entire dataset at this time. 

\begin{table}[tbp]
    \centering
    \begin{tabular}{c|ccccc}
        & Precision~\upward & Recall~\upward & F1~\upward & p-value~\downward \\ \hline
        Random & 10.64 & 15.53 & 12.63  &  0.56 \\
        QA-Temp & 31.83 & 22.96 & 26.67 & 0.00 \\ 
        QA-Guide & \textbf{31.89} & 23.08 & 26.78 & 0.00 \\ 
        QA-Trig & 28.03 & \textbf{26.26} & 27.12 & 0.00\\ 
        \rowcolor{LightCyan} QA-Trig-Plus & 30.05 & 24.80 & \textbf{27.17} & 0.00
    \end{tabular}
    \caption{Zero-shot learning for argument detection. We report Precision, Recall, F1, and \(p\)-values. The arguments are predicted based on ground-truth event labels and trigger word annotations. }
    \label{tab:zero_shot_arg_results}
\end{table}

\begin{table*}[tbp]
    \small
    \centering
    \begin{tabular}{p{0.43\textwidth}|c|c|c|c}
        Sentence & Event, Argument, \& Answer & Query Type & Prediction & Result \\ \hline
        \multirow{4}{0.43\textwidth}{\textbf{What is the organization he said that is going to start?} ``Prostitution is completely discriminalised in Sydney and we are going to build a monster,'' he said.}  & \multirow{4}*{\shortstack{Start-Organization, \\ Organization, \\ ``a monster''}}  & QA-Temp & ``a monster'' & Correct \\ 
        & & QA-Guide & None & Wrong \\
        & & QA-Trigger & ``a monster'' & Correct \\ 
        & & Custom & ``Universal \(\ldots\) assets'' & Partial correct \\\hline

        \multirow{4}{0.43\textwidth}{\textbf{Who started a new job in London's financial world?} Former senior banker Callum McCarthy begins what is one of the most important jobs in London 's financial world in September , when incumbent Howard Davies steps down.  }  & \multirow{4}*{\shortstack{Start-Position, \\ Person, \\ ```Former \(\ldots\) McCarthy''}}  & QA-Temp & ``Former \(\ldots\) McCarthy'' & Correct \\ 
        & & QA-Guide & ``Former \(\ldots\) McCarthy'' & Correct \\
        & & QA-Trigger & None & Wrong \\ 
        & & Custom & ``Callum McCarthy'' & Partial correct \\\hline

        \multirow{4}{0.43\textwidth}{\textbf{What are the entertainment assets Vivendi tries to sell by the end of the year?} Vivendi confirmed that it planned to shed its entertainment assets by the end of the year, including its famed Universal movie studio and television assets.}  & \multirow{4}*{\shortstack{Transfer-Ownership, \\ Artifact, \\ ``its famed\(\ldots\) studio''}}  & QA-Temp & None & Wrong \\ 
        & & QA-Guide & None & Wrong \\
        & & QA-Trigger & None & Wrong \\ 
        & & Custom & ``Universal \(\ldots\) assets'' & Partial correct \\\hline

        \multirow{4}{0.43\textwidth}{\textbf{When did the meeting that Jean-Rene Fourtou participated in take place?} Chief executive Jean - Rene Fourtou told shareholders at the group 's annual general meeting Tuesday that \hide{the sale of Vivendi Universal Entertainment was a major goal for 2003 , and that} negotiations were already under way.}  & \multirow{4}*{\shortstack{Meet, \\ Time-Within, \\ ``Tuesday''}}  & QA-Temp & None & Wrong \\ 
        & & QA-Guide & None & Wrong \\
        & & QA-Trigger & None & Wrong \\ 
        & & Custom & ``Tuesday'' & Correct \\\hline
    \end{tabular}
    \hide{
    \begin{tabular}{l|c}
        Index & Custom Query \\ \hline
        1 & \\\hline
        2 & What are the entertainment assets Vivendi are trying to sell by the end of the year? \\\hline
        3 & When did the meeting that Jean-Rene Fourtou participated in take place \\  
    \end{tabular}
    }
    \caption{Zero-shot learning for question answering examples. The bold first sentence is the ``Custom'' query that human manually wrote given the context. We list predictions by three generated queries and the custom query. Some non-essential parts are removed from sentences due to space limitations.  }
    \label{tab:exp:arg_examples}
\end{table*}

\subsubsection{Few-shot learning}
\label{sec:arg:sup}
To the best of our knowledge, we are the first to do few-shot argument detection without using external semantic role labeling tools. We show our results in Table~\ref{tab:few_shot_arg_results}.

In Table~\ref{tab:few_shot_arg_results}, we report rgument prediction scores based on both event predictions using TE model from Section~\ref{sec:event:sup} and on ground-truth event labels. 
Errors in event predictions would propagate in the first scenario. 
We see that the hand-written query based on guides perform a lot better, especially when \(K\) is low, than template-based queries. 

% QA-Temp, QA-Guide, QA-Trig, QA-Trig-Plus

\begin{table*}[htbp]
    \centering
    \begin{tabular}{c|c|ccccccccc}
    Events & \(K\)-shot & \(K=1\) & \(K=3\) & \(K=5\) & \(K=7\) & \(K=9\) \\ \hline
     \multirow{2}*{Predicted}  & QA-Temp & 1.35 \std{2.34} & 37.27 \std{2.28} & 44.08 \std{0.74} & 44.74 \std{2.87} & 46.59 \std{0.94} \\
      &  QA-Guide & 19.58 \std{2.45} & 41.38 \std{1.61} & 45.41 \std{1.48} & 44.92 \std{1.12} & 46.86 \std{0.74} \\ \hline
    \multirow{2}*{Ground Truth} & QA-Temp & 1.62 \std{2.81} & 45.49 \std{2.46} & 52.49 \std{1.32} & 54.44 \std{3.19} & 56.15 \std{1.13} \\
       &  QA-Guide & 22.67 \std{3.03} & 50.97 \std{2.42} & 55.48 \std{1.68} & 54.39 \std{0.85} & 57.66 \std{1.23} \\ 
      % &  QA-Trig &  0.00 \std{0.00} & 57.44 \std{1.19} & 61.46 \std{1.05} & 62.65 \std{0.41} & 63.49 \std{6.94} \\
       %&  QA-Trig-Plus & \\
\end{tabular}
    \caption{Results on few-shot argument detection. The first part predicts and compute metrics based on predicted events with the best text entailment model trained in Section~\ref{sec:method:event}. The second part is based on ground-truth event labels. }
    \label{tab:few_shot_arg_results}
\end{table*}

\subsubsection{Fully Supervised}
For fully supervised argument detection, we compare our approach to DMCNN~\cite{chen2015event} and VERB-QA~\cite{du2020event} described in Section~\ref{sec:event:sup}.

VERB-QA~\cite{du2020event} curates the argument with a trigger-word based QA framework. Specifically, their template is: 
``
\texttt{[Wh-word]} is the \texttt{Argument} in \texttt{Trigger}?
''
where ``Wh-word'' means interrogative words such as ``What'', ``Who'' and ``When'', and ``Trigger'' is the detected trigger word. This is essentially equivalent to our QA-Trig.
%VERB-QA~\cite{du2020event} achieved the best result among baselines. 
In ~\ref{tab:arg_sup} we report the experiment results. Our method with 

\begin{table}
    \small
    \begin{tabular}{l|l|ccc}
      Events & Method & Precision & Recall & F1\\ \hline
       \multirow{4}*{Pred.} &  DMCNN & \textbf{62.2} & 46.9 & 53.5 \\ 
        & VERB-QA & 56.77  & 50.24 & 53.31 \\
       & QA-Temp &  56.24~\std{1.21} & \textbf{52.14}~\std{3.52} & 54.01~\std{2.13} \\ 
        \rowcolor{LightCyan}& QA-Guide & 57.69~\std{0.45} & 51.90~\std{3.39} & \textbf{54.61}~\std{1.67} \\ 
        \hline
        & QA-Temp & 71.06~\std{1.58} & \textbf{65.84}~\std{2.14} & 68.25~\std{0.98} \\ 
        \rowcolor{LightCyan}
         \multirow{1}{*}{G. Truth} & QA-Guide & \textbf{72.20}~\std{2.45} & 65.79~\std{3.09} & \textbf{68.79}~\std{0.57}
    \end{tabular}
    \caption{Fully supervised learning for argument detection. Similar to in few-shot setting, we report our scores on both ground-truth and predicted event labels. }
    \label{tab:arg_sup}
\end{table}

\subsubsection{Discussions}
\label{sec:arg:ablation}
% in zero-shot, yes, because it provides only source of relatedness to the data. 
% in supervised, less so.
% considering error propagation, not that much. 
%\paragraph{On query formulations.}
%One problem with generated queries, either from templates or descriptions from guidance, is that they can hardly relate to the context of the sentence and other arguments, which can be crucial for the model to accurately pinpoint the answer. 

\paragraph{Does trigger word help?}
The potential drawback of event detection without triggers is that it might lose the trigger information that could be important for subsequent argument detection tasks. 
In Section~\ref{sec:arg:zero_shot}, we see that questions with trigger words perform slightly better than questions with only event and argument names. 
Since trigger words contain the only clue in the query about the context of the sentence, it is indeed reasonable that questions with triggers should perform better.

However, this is based on golden trigger words. In real application, predicting trigger words tend to propagate more errors than predicting sentence-level event labels.
Hence, in supervised setting, we see that methods based on sentence-level event predictions (QA-Temp, QA-Guide) perform slightly better than VERB-QA, which predicts arguments based on predicted trigger words and questions constructed with them. 
We may conclude that trigger words are non-essential to both event detection and argument detection. 

\section{Related Work}

In this section, we review related works from closely relevant fields. First, we review the most relevant reading comprehension framework and applications to NLP tasks (Section~\ref{sec:related:rc}). 
Second, we review relevant papers from event extraction,  in both standard supervised setting (Section~\ref{sec:related:ed}) and few-shot learning setting (Section~\ref{sec:related:fsl}). 
At last, since our method implicitly treat \bert as a knowledge base, we review relevant papers in Section~\ref{sec:related:knowledge}. 

\subsection{Reading Comprehension Frameworks}
\label{sec:related:rc}
We first review methods that reframe various NLP tasks as machine reading comprehension. 
Most trending natural language processing tasks can be formulated as reading comprehension tasks. \cite{levy2017zero_shot_re} used pre-\bert model for question answering for zero-shot relation extraction. 
Later on, more works formulated tasks reading comprehension, such as \cite{keskar2019qa_sentence} for sentence classification, \cite{li2019unified_mrc_ner} for NER, \cite{obamuyide2018zero_shot_re_entail} for relation extraction with entailment prediction, \cite{wu2020corefqa-query} for coreference resolution, and \cite{wu2019zero_qa_slot_filling} for slot filling. \cite{mccann2018natural_qa_multitask} unified 10 tasks into one question answering based multi-task learning framework. \cite{das2018building_kg} used reading comprehension as a tool to build knowledge graphs. \cite{gardner2019question_useful} probed whether or when the format of question answering is useful.
\cite{chen2019reading_manual_ee} designed a pipeline to do zero-shot event detection by reading information from the annotation guide. 

Formulating tasks as machine reading comprehension has one key advantage, that it is able to better utilize knowledge in modern language models, such as \bert~\cite{devlin2018bert_}, ideally improving label efficiency. 
For example, among above mentioned works, \cite{levy2017zero_shot_re,obamuyide2018zero_shot_re_entail,wu2019zero_qa_slot_filling,chen2019reading_manual_ee} all employed reading comprehension as a tool for zero- or few-shot learning. 

\cite{du2020event} is closley related to our work in that they also developed a reading comprehension framework for event detection. The key difference is that, they used a QA framework for event trigger detection and classification. As shown in Section~\ref{sec:event_detection}, this isn't necessarily the most natural way to utilize \bert. Indeed, \cite{du2020event} found that the most effective question form is a single word ``verb'', essentailly not a question and only hinting the model that verbs are more possible trigger words, which is not the ideal kind of knowledge one expects to extract from \bert.

\subsection{Event Detection}
\label{sec:related:ed}
The early attempts to solve event detection rely on hand-crafted features ~\cite{li2013joint} and probabilisitc rules ~\cite{liu2016probabilistic}.
More recent works utilize neural networks for learning salient representations for event detection, such as CNN~\cite{chen2015event,nguyen2015event,nguyen2015event}, RNN~\cite{nguyen2016joint,duan2017exploiting,hong2018self,sha2019jointly}, and GNN~\cite{liu2018jointly,nguyen2018graph,yan2019event}. 

Unlike other sequence tagging problems, such as NER, event extraction relies more heavily on the structural information in the sentence. Hence, one way to predict events and arguments jointly is by utilizing dependency parse~\cite{nguyen2018graph,sha2019jointly,liu2018jointly,yan2019event}. 
Many researchers exploit the hierachical information as well~\cite{liao2010using,yang2018dcfee,duan2017exploiting,yang2016hierarchical,mehta2019event}.

Similarly to our setting, \cite{liu2019event} considers event detection a sentence classification task, citing that the subjectivity in tagging event triggers can harm model performance and the knowledge of triggers is ``non-essential to the task''~\cite{liu2019event}. However, they did not further extend the framework to include argument detection.
To the best of our knowledge, we are the first to implement argument detection without relying on trigger information.

\subsection{Few-shot Learning}
\label{sec:related:fsl}
There exists two approaches to few-shot event detection.
The first is based on defining prototype vectors for events~\cite{peng2016event,huang2017zero}.
They both extract a graphical structure for each candidate trigger words with external resources, including semantic role labeling~\cite{peng2016event} and abstract meaning representation~\cite{huang2017zero}. 
SRL and AMR themselves are tantamount to argument detection already, so their works can be viewed as event detectoin based on given arguments. Since we focus on doing both event detection and argument detection from scratch, their methods aren't directly comparable to ours. It would be a promising direction to incorporate their structural knowledge into our framework. 

The second approach that has been applied to few-shot event detection is meta-learning in ~\cite{deng2020meta}. They have built their own data based on ACE05~\cite{doddington2004automatic}. Since we do not have access to their customized data or their implementation, we could not compare our method with theirs. 

\subsection{\bert as a Knowledge Base}
\label{sec:related:knowledge}
There have been works that investigate whether Bert is a knowledge base itself.
The hypothesis is that Bert, while learning from massive corpora, could memorize factual and commonsense knowledge about the world. ~\cite{petroni2019language,roberts2020much}.
\cite{petroni2019language}, Bert might have inferred right relations, without having the right understanding, but instead based on ``learned associations of objects with subjects from co-occurrence patterns''. Still, the work of \cite{bouraoui2019inducing} shows that fine-tuned Bert model can consistently outperform simple word-vector-based models in inferring relations. \cite{roberts2020much} extends to use Bert to answer more complex natural language queries, instead of traditional triplet queries, and shows that Bert outperforms open-domain retrieval baselines by a large margin.
\section{Conclusion}
We propose a reading comprehension framework for event extraction tasks. 
We design a textual entailment based method for event detection and question answering for argument detection. 
Experiment findings suggest our framework can effectively distill knowledge from \bert as well as guide the model with semantic information, achieving state-of-the-art results on few-shot and supervised settings.

Our experiemnts also suggest several promising reseach directions. We summarize them here. 
It is clear that while the current framework with mostly template-based queries can achieve superior performance already, 
the queries are not ideal since they do not relate to actual sentence context and the casual relation between distinctive events and arguments. 
One promising direction is generating queries that are 1) more flexible and authentic, 2) relevant to input sentence's context, and 3) reveals causal relations between events and arguments. We believe this would enable a more label efficient and robust zero-shot and few-shot learning framework.

In our current framework and, indeed, all existing reading comprehension frameworks to our best knowledge, 
one must construct multiple queries for one input instance for complete classification results. 
This could be burdensome when there's a large number of queries to be constructed, usually a result of large number of label types. 
One problem is how to do so efficiently by enabling information sharing between different queries. 

Experiments show that the current \bert model cannot learn efficiently from long event descriptions. A siginificant advancement in language modeling would be enabling it for zero- or few-shot learning with only a few descriptions and annotation guides. 

We manually select reading comprehension tasks for two event extraction tasks. 
%A general reading comprehension framework could 
A key ingredient to a more general reading comprehension solution to NLP tasks is a principle to measure the transferibility between tasks.

\bibliographystyle{acm}
\bibliography{nlp,events,stats}

\begin{thebibliography}{10}

\bibitem{bouraoui2019inducing}
{\sc Bouraoui, Z., Camacho-Collados, J., and Schockaert, S.}
\newblock Inducing relational knowledge from bert.
\newblock {\em arXiv preprint arXiv:1911.12753\/} (2019).

\bibitem{chen2019reading_manual_ee}
{\sc Chen, Y., Chen, T., Ebner, S., and Van~Durme, B.}
\newblock Reading the manual: Event extraction as definition comprehension.
\newblock {\em arXiv preprint arXiv:1912.01586\/} (2019).

\bibitem{chen2017automatically}
{\sc Chen, Y., Liu, S., Zhang, X., Liu, K., and Zhao, J.}
\newblock Automatically labeled data generation for large scale event
  extraction.
\newblock In {\em Proceedings of the 55th Annual Meeting of the Association for
  Computational Linguistics (Volume 1: Long Papers)\/} (2017), pp.~409--419.

\bibitem{chen2015event}
{\sc Chen, Y., Xu, L., Liu, K., Zeng, D., and Zhao, J.}
\newblock Event extraction via dynamic multi-pooling convolutional neural
  networks.
\newblock In {\em Proceedings of the 53rd Annual Meeting of the Association for
  Computational Linguistics and the 7th International Joint Conference on
  Natural Language Processing (Volume 1: Long Papers)\/} (Beijing, China, July
  2015), Association for Computational Linguistics, pp.~167--176.

\bibitem{ace05guide}
{\sc Consortium, L.~D.}
\newblock Ace (automatic content extraction) english annotation guidelines for
  events.
\newblock
  \url{https://www.ldc.upenn.edu/sites/www.ldc.upenn.edu/files/english-events-guidelines-v5.4.3.pdf}.

\bibitem{das2018building_kg}
{\sc Das, R., Munkhdalai, T., Yuan, X., Trischler, A., and McCallum, A.}
\newblock Building dynamic knowledge graphs from text using machine reading
  comprehension.
\newblock {\em arXiv preprint arXiv:1810.05682\/} (2018).

\bibitem{deng2020meta}
{\sc Deng, S., Zhang, N., Kang, J., Zhang, Y., Zhang, W., and Chen, H.}
\newblock Meta-learning with dynamic-memory-based prototypical network for
  few-shot event detection.
\newblock In {\em Proceedings of the 13th International Conference on Web
  Search and Data Mining\/} (2020), pp.~151--159.

\bibitem{devlin2018bert_}
{\sc Devlin, J., Chang, M.-W., Lee, K., and Toutanova, K.}
\newblock Bert: Pre-training of deep bidirectional transformers for language
  understanding.
\newblock {\em arXiv preprint arXiv:1810.04805\/} (2018).

\bibitem{doddington2004automatic}
{\sc Doddington, G.~R., Mitchell, A., Przybocki, M.~A., Ramshaw, L.~A.,
  Strassel, S.~M., and Weischedel, R.~M.}
\newblock The automatic content extraction (ace) program-tasks, data, and
  evaluation.
\newblock In {\em Lrec\/} (2004), vol.~2, Lisbon, pp.~837--840.

\bibitem{du2020event}
{\sc Du, X., and Cardie, C.}
\newblock Event extraction by answering (almost) natural questions.
\newblock {\em arXiv preprint arXiv:2004.13625\/} (2020).

\bibitem{duan2017exploiting}
{\sc Duan, S., He, R., and Zhao, W.}
\newblock Exploiting document level information to improve event detection via
  recurrent neural networks.
\newblock In {\em Proceedings of the Eighth International Joint Conference on
  Natural Language Processing (Volume 1: Long Papers)\/} (Taipei, Taiwan, Nov.
  2017), Asian Federation of Natural Language Processing, pp.~352--361.

\bibitem{gardner2019question_useful}
{\sc Gardner, M., Berant, J., Hajishirzi, H., Talmor, A., and Min, S.}
\newblock Question answering is a format; when is it useful?
\newblock {\em arXiv preprint arXiv:1909.11291\/} (2019).

\bibitem{hodges1958significance}
{\sc Hodges, J.~L.}
\newblock The significance probability of the smirnov two-sample test.
\newblock {\em Arkiv f{\"o}r Matematik 3}, 5 (1958), 469--486.

\bibitem{hong2011using}
{\sc Hong, Y., Zhang, J., Ma, B., Yao, J., Zhou, G., and Zhu, Q.}
\newblock Using cross-entity inference to improve event extraction.
\newblock In {\em Proceedings of the 49th Annual Meeting of the Association for
  Computational Linguistics: Human Language Technologies-Volume 1\/} (2011),
  Association for Computational Linguistics, pp.~1127--1136.

\bibitem{hong2018self}
{\sc Hong, Y., Zhou, W., Zhang, J., Zhou, G., and Zhu, Q.}
\newblock Self-regulation: Employing a generative adversarial network to
  improve event detection.
\newblock In {\em Proceedings of the 56th Annual Meeting of the Association for
  Computational Linguistics (Volume 1: Long Papers)\/} (Melbourne, Australia,
  July 2018), Association for Computational Linguistics, pp.~515--526.

\bibitem{huang2017zero}
{\sc Huang, L., Ji, H., Cho, K., and Voss, C.~R.}
\newblock Zero-shot transfer learning for event extraction.
\newblock {\em arXiv preprint arXiv:1707.01066\/} (2017).

\bibitem{keskar2019qa_sentence}
{\sc Keskar, N.~S., McCann, B., Xiong, C., and Socher, R.}
\newblock Unifying question answering, text classification, and regression via
  span extraction.
\newblock {\em arXiv preprint arXiv:1904.09286\/} (2019).

\bibitem{levy2017zero_shot_re}
{\sc Levy, O., Seo, M., Choi, E., and Zettlemoyer, L.}
\newblock Zero-shot relation extraction via reading comprehension.
\newblock {\em arXiv preprint arXiv:1706.04115\/} (2017).

\bibitem{li2013joint}
{\sc Li, Q., Ji, H., and Huang, L.}
\newblock Joint event extraction via structured prediction with global
  features.
\newblock In {\em Proceedings of the 51st Annual Meeting of the Association for
  Computational Linguistics (Volume 1: Long Papers)\/} (2013), pp.~73--82.

\bibitem{li2019unified_mrc_ner}
{\sc Li, X., Feng, J., Meng, Y., Han, Q., Wu, F., and Li, J.}
\newblock A unified mrc framework for named entity recognition.
\newblock {\em arXiv preprint arXiv:1910.11476\/} (2019).

\bibitem{liao2010using}
{\sc Liao, S., and Grishman, R.}
\newblock Using document level cross-event inference to improve event
  extraction.
\newblock In {\em Proceedings of the 48th Annual Meeting of the Association for
  Computational Linguistics\/} (2010), Association for Computational
  Linguistics, pp.~789--797.

\bibitem{liu2019event}
{\sc Liu, S., Li, Y., Zhang, F., Yang, T., and Zhou, X.}
\newblock Event detection without triggers.
\newblock In {\em Proceedings of the 2019 Conference of the North American
  Chapter of the Association for Computational Linguistics: Human Language
  Technologies, Volume 1 (Long and Short Papers)\/} (2019), pp.~735--744.

\bibitem{liu2016probabilistic}
{\sc Liu, S., Liu, K., He, S., and Zhao, J.}
\newblock A probabilistic soft logic based approach to exploiting latent and
  global information in event classification.
\newblock In {\em Thirtieth AAAI Conference on Artificial Intelligence\/}
  (2016).

\bibitem{liu2018jointly}
{\sc Liu, X., Luo, Z., and Huang, H.}
\newblock Jointly multiple events extraction via attention-based graph
  information aggregation.
\newblock {\em arXiv preprint arXiv:1809.09078\/} (2018).

\bibitem{lu2019distilling}
{\sc Lu, Y., Lin, H., Han, X., and Sun, L.}
\newblock Distilling discrimination and generalization knowledge for event
  detection via delta-representation learning.
\newblock In {\em Proceedings of the 57th Annual Meeting of the Association for
  Computational Linguistics\/} (2019), pp.~4366--4376.

\bibitem{massey1951kolmogorov}
{\sc Massey~Jr, F.~J.}
\newblock The kolmogorov-smirnov test for goodness of fit.
\newblock {\em Journal of the American statistical Association 46}, 253 (1951),
  68--78.

\bibitem{mccann2018natural_qa_multitask}
{\sc McCann, B., Keskar, N.~S., Xiong, C., and Socher, R.}
\newblock The natural language decathlon: Multitask learning as question
  answering.
\newblock {\em arXiv preprint arXiv:1806.08730\/} (2018).

\bibitem{mehta2019event}
{\sc Mehta, S., Islam, M.~R., Rangwala, H., and Ramakrishnan, N.}
\newblock Event detection using hierarchical multi-aspect attention.
\newblock In {\em The World Wide Web Conference\/} (New York, NY, USA, 2019),
  WWW '19, ACM, pp.~3079--3085.

\bibitem{nguyen2018graph}
{\sc Nguyen, T., and Grishman, R.}
\newblock Graph convolutional networks with argument-aware pooling for event
  detection.

\bibitem{nguyen2016joint}
{\sc Nguyen, T.~H., Cho, K., and Grishman, R.}
\newblock Joint event extraction via recurrent neural networks.
\newblock In {\em HLT-NAACL\/} (2016), pp.~300--309.

\bibitem{nguyen2015event}
{\sc Nguyen, T.~H., and Grishman, R.}
\newblock Event detection and domain adaptation with convolutional neural
  networks.
\newblock In {\em Proceedings of the 53rd Annual Meeting of the Association for
  Computational Linguistics and the 7th International Joint Conference on
  Natural Language Processing (Volume 2: Short Papers)\/} (2015), pp.~365--371.

\bibitem{obamuyide2018zero_shot_re_entail}
{\sc Obamuyide, A., and Vlachos, A.}
\newblock Zero-shot relation classification as textual entailment.
\newblock In {\em Proceedings of the First Workshop on Fact Extraction and
  VERification (FEVER)\/} (2018), pp.~72--78.

\bibitem{orr2018event}
{\sc Orr, J.~W., Tadepalli, P., and Fern, X.}
\newblock Event detection with neural networks: A rigorous empirical
  evaluation.
\newblock {\em arXiv preprint arXiv:1808.08504\/} (2018).

\bibitem{peng2016event}
{\sc Peng, H., Song, Y., and Roth, D.}
\newblock Event detection and co-reference with minimal supervision.
\newblock In {\em Proceedings of the 2016 conference on empirical methods in
  natural language processing\/} (2016), pp.~392--402.

\bibitem{petroni2019language}
{\sc Petroni, F., Rockt{\"a}schel, T., Lewis, P., Bakhtin, A., Wu, Y., Miller,
  A.~H., and Riedel, S.}
\newblock Language models as knowledge bases?
\newblock {\em arXiv preprint arXiv:1909.01066\/} (2019).

\bibitem{rajpurkar2016squad}
{\sc Rajpurkar, P., Zhang, J., Lopyrev, K., and Liang, P.}
\newblock Squad: 100,000+ questions for machine comprehension of text.
\newblock {\em arXiv preprint arXiv:1606.05250\/} (2016).

\bibitem{roberts2020much}
{\sc Roberts, A., Raffel, C., and Shazeer, N.}
\newblock How much knowledge can you pack into the parameters of a language
  model?
\newblock {\em arXiv preprint arXiv:2002.08910\/} (2020).

\bibitem{schuster2012wordpiece}
{\sc Schuster, M., and Nakajima, K.}
\newblock Japanese and korean voice search.
\newblock In {\em 2012 IEEE International Conference on Acoustics, Speech and
  Signal Processing (ICASSP)\/} (2012), IEEE, pp.~5149--5152.

\bibitem{sha2019jointly}
{\sc Sha, L., Qian, F., Chang, B., and Sui, Z.}
\newblock Jointly extracting event triggers and arguments by dependency-bridge
  rnn and tensor-based argument interaction.

\bibitem{wang2019adversarial}
{\sc Wang, X., Han, X., Liu, Z., Sun, M., and Li, P.}
\newblock Adversarial training for weakly supervised event detection.
\newblock In {\em Proceedings of the 2019 Conference of the North American
  Chapter of the Association for Computational Linguistics: Human Language
  Technologies, Volume 1 (Long and Short Papers)\/} (2019), pp.~998--1008.

\bibitem{mnli}
{\sc Williams, A., Nangia, N., and Bowman, S.}
\newblock A broad-coverage challenge corpus for sentence understanding through
  inference.
\newblock In {\em Proceedings of the 2018 Conference of the North American
  Chapter of the Association for Computational Linguistics: Human Language
  Technologies, Volume 1 (Long Papers)\/} (2018), Association for Computational
  Linguistics, pp.~1112--1122.

\bibitem{wu2019zero_qa_slot_filling}
{\sc Wu, T., Wang, M., Gao, H., Qi, G., and Li, W.}
\newblock Zero-shot slot filling via latent question representation and reading
  comprehension.
\newblock In {\em Pacific Rim International Conference on Artificial
  Intelligence\/} (2019), Springer, pp.~123--136.

\bibitem{wu2020corefqa-query}
{\sc Wu, W., Wang, F., Yuan, A., Wu, F., and Li, J.}
\newblock Corefqa: Coreference resolution as query-based span prediction.
\newblock In {\em Proceedings of the 58th Annual Meeting of the Association for
  Computational Linguistics\/} (2020), pp.~6953--6963.

\bibitem{yan2019event}
{\sc Yan, H., Jin, X., Meng, X., Guo, J., and Cheng, X.}
\newblock Event detection with multi-order graph convolution and aggregated
  attention.
\newblock In {\em Proceedings of the 2019 Conference on Empirical Methods in
  Natural Language Processing and the 9th International Joint Conference on
  Natural Language Processing (EMNLP-IJCNLP)\/} (2019), pp.~5770--5774.

\bibitem{yang2018dcfee}
{\sc Yang, H., Chen, Y., Liu, K., Xiao, Y., and Zhao, J.}
\newblock Dcfee: A document-level chinese financial event extraction system
  based on automatically labeled training data.
\newblock {\em ACL 2018\/} (2018), 50.

\bibitem{yang2019exploring}
{\sc Yang, S., Feng, D., Qiao, L., Kan, Z., and Li, D.}
\newblock Exploring pre-trained language models for event extraction and
  generation.
\newblock In {\em Proceedings of the 57th Annual Meeting of the Association for
  Computational Linguistics\/} (2019), pp.~5284--5294.

\bibitem{yang2016hierarchical}
{\sc Yang, Z., Yang, D., Dyer, C., He, X., Smola, A.~J., and Hovy, E.~H.}
\newblock Hierarchical attention networks for document classification.
\newblock In {\em HLT-NAACL\/} (2016).

\end{thebibliography}

\end{document}